\documentclass[conference]{IEEEtran}
\IEEEoverridecommandlockouts

\usepackage{cite}
\usepackage{amsmath,amssymb,amsfonts}%
\usepackage{algorithmic}
\usepackage{graphicx}
\usepackage{colortbl}
\usepackage{textcomp}
\usepackage{xcolor}
\usepackage{booktabs}
\usepackage{hyperref}
\usepackage{svg}
\usepackage[capitalize]{cleveref}
\usepackage{fancyhdr}

\def\BibTeX{{\rm B\kern-.05em{\sc i\kern-.025em b}\kern-.08em
    T\kern-.1667em\lower.7ex\hbox{E}\kern-.125emX}}
\begin{document}

\title{Benchmarking Music Generation Models and Metrics via Human Preference Studies
\thanks{${^*}$Equal contribution.}
}

\author{\IEEEauthorblockN{Florian Grötschla${^*}$}
\IEEEauthorblockA{%
\textit{ETH Zurich}\\
fgroetschla@ethz.ch}
\and
\IEEEauthorblockN{Ahmet Solak${^*}$}
\IEEEauthorblockA{%
\textit{ETH Zurich}\\
asolak@ethz.ch}
\and
\IEEEauthorblockN{Luca A. Lanzendörfer}
\IEEEauthorblockA{%
\textit{ETH Zurich}\\
lanzendoerfer@ethz.ch}
\and
\IEEEauthorblockN{Roger Wattenhofer}
\IEEEauthorblockA{%
\textit{ETH Zurich}\\
wattenhofer@ethz.ch}
}

\maketitle

\def\@IEEEpubidpullup{8\baselineskip}
\makeatother
\IEEEpubid{
    \begin{minipage}{\textwidth}
        \vspace{2cm}
        \centering
        \footnotesize
        © 2025 IEEE. Personal use of this material is permitted. This work has been accepted to \textit{ICASSP 2025}. %
    Permission from IEEE must be obtained for all other uses, in any current or future media, including reprinting/%
    republishing this material for advertising or promotional purposes, creating new collective works, for resale or redistribution to servers or lists, or reuse of any copyrighted component of this work in other works.\\
    The final version is published by IEEE and is available at: \texttt{https://doi.org/10.1109/ICASSP49660.2025.10887745}
    \end{minipage}
}

\begin{abstract}
Recent advancements have brought generated music closer to human-created compositions, yet evaluating these models remains challenging. While human preference is the gold standard for assessing quality, translating these subjective judgments into objective metrics, particularly for text-audio alignment and music quality, has proven difficult. In this work, we generate 6k songs using 12 state-of-the-art models and conduct a survey of 15k pairwise audio comparisons with 2.5k human participants to evaluate the correlation between human preferences and widely used metrics. To the best of our knowledge, this work is the first to rank current state-of-the-art music generation models and metrics based on human preference. To further the field of subjective metric evaluation, we provide open access to our dataset of generated music and human evaluations.
\end{abstract}


\section{Introduction}

The field of AI-generated music has witnessed unprecedented progress, with recent models producing compositions that are becoming increasingly indistinguishable from those created by humans. As these advancements continue, the evaluation of AI-generated music becomes even more relevant. While human preference remains the gold standard for assessing the quality and effectiveness of these models, translating these subjective judgments into reliable objective metrics remains an open challenge.
Efforts to bridge this gap have largely focused on two key aspects: (1) The quality of alignment between the text prompt and the audio and (2) the overall quality of the generated music. Despite the development of various objective metrics to evaluate these aspects, their effectiveness in capturing human preferences remains uncertain. In this context, our study seeks to assess the correlation between human evaluations and widely used objective metrics in music generation.

We generate 6,000 songs using 12 state-of-the-art music generative models and conduct a large-scale survey involving 15,600 pairwise audio comparisons with more than 2,500 human participants. These comparisons were designed to evaluate text-audio alignment and music preference from a human perspective and to compare these evaluations with existing objective metrics. Our results show that certain metrics align better with human judgment than others, allowing us to create a comprehensive ranking.  

To facilitate further research in this area, we make both the generated music dataset and the human evaluation dataset publicly available. Our contributions provide a foundation for future work aimed at improving the evaluation of AI-generated music and offer a testing ground and benchmark for new metrics that align more closely with human perception. 

\begin{figure}[t!]
\centerline{\includegraphics[width=\linewidth]{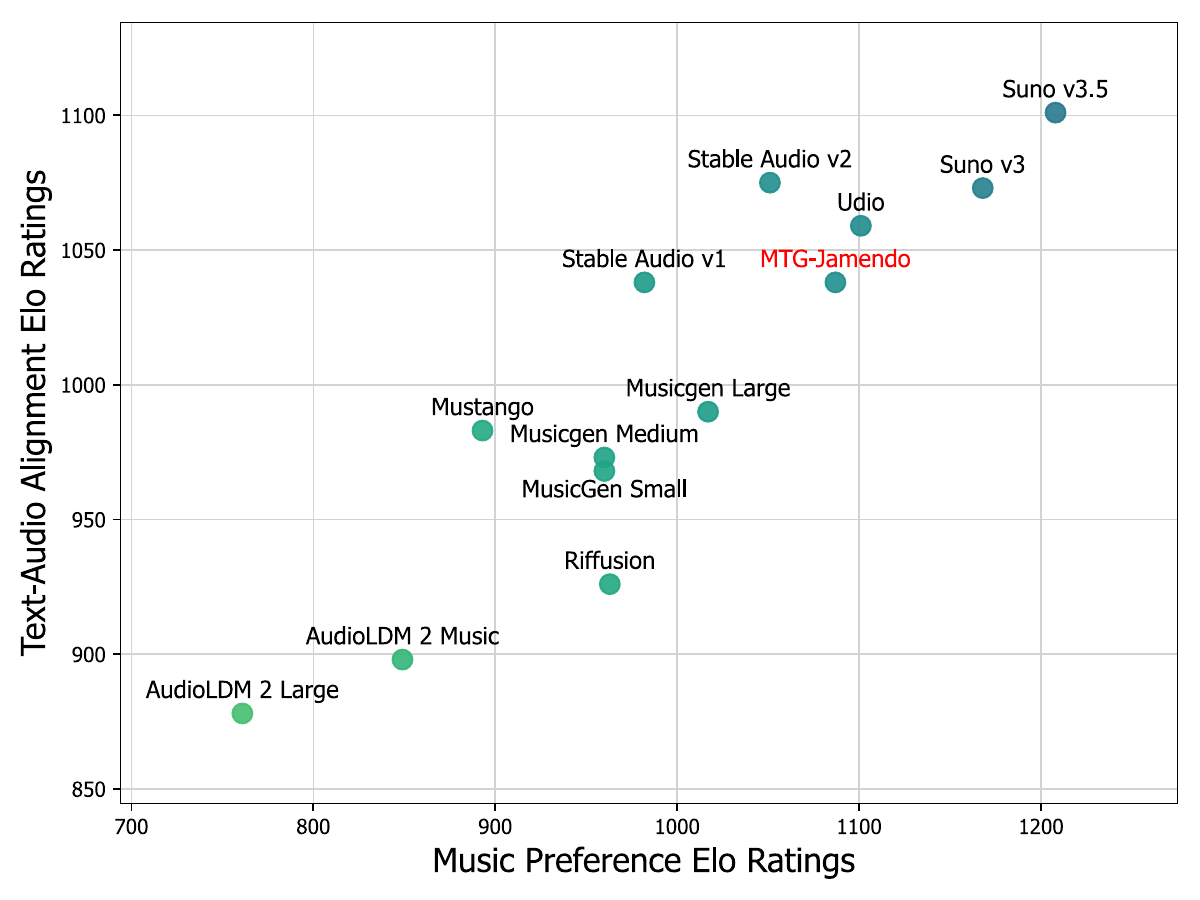}}
\caption{Elo ratings for all music generation models in the music preference and text-audio alignment human evaluation experiments. The reference dataset is shown in red.}
\label{fig_wins_scatterplot}
\end{figure}

\begin{figure*}
\centerline{\includegraphics[width=\textwidth]{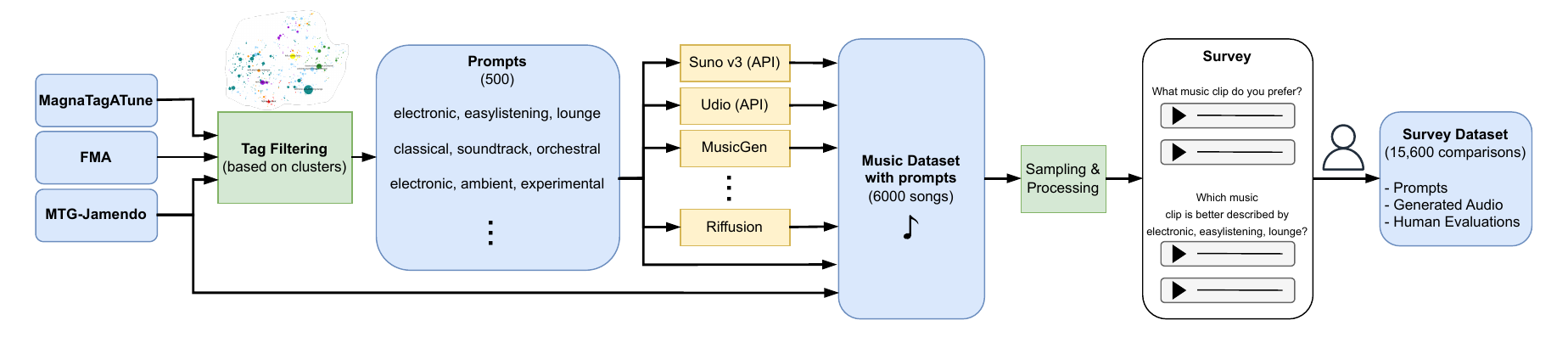}}
\caption{Overview of the tag selection, music generation, and study. We extract common and diverse tag combinations from MTG-Jamendo and use them to generate 6k music snippets with a series of models. We sample from this corpus to generate a survey dataset and let human evaluators compare the samples based on text-audio alignment and music quality.}
\label{fig_survey}
\end{figure*}

Our contributions can be summarized as follows: 
\begin{itemize} 
    \item We conduct a large-scale survey of over 15k audio comparisons with 2.5k human participants to understand human preference for text-audio alignment and overall music preference. To this end, we create a dataset of 6k generated songs using 12 state-of-the-art music generation models. 
    \item We analyze the relationship between human preference and existing metrics used in the domain of music generation. We find significant differences in their alignment and identify the metrics that reflect human perception the best.
    \item We open-source the tags, prompts, and generated songs, as well as the human responses to the survey, to enable further testing with new metrics and models.\footnote{Available at \url{https://huggingface.co/datasets/disco-eth/AIME}}
\end{itemize}

\section{Related Work}

\label{sec:format}

The Frechet Audio Distance (FAD)~\cite{kilgour2018fr} metric serves as a quantitative measure of the perceptual quality of generated audio. FAD is computed on audio embeddings and measures the distribution similarity between a set of generated audio and ground-truth audio. FAD is commonly used on audio embeddings generated with VGGish~\cite{hershey2017cnn}. %
MusicLM~\cite{agostinelli2023musiclm}, utilized the FAD metric 
to demonstrate superior audio and music quality in comparison with previous models~\cite{forsgren2022riffusion}. 
Although FAD is a key metric for evaluating the perceptual quality of AI-generated music, the CLAP model \cite{elizalde2023clap,elizalde2024natural,wu2023large} is widely used to measure the alignment between generated audio and text prompt. 

Given the variety of objective metrics available, there have been efforts to assess their reliability and effectiveness, particularly in evaluating perceptual qualities such as audio and music quality. There has also been work done comparing the scores produced by objective metrics with those obtained from listening studies involving neural network-generated audio \cite{vinay2022evaluating}. Those findings suggested that current objective metrics might not fully capture the perceptual quality of audio. More recent research has explored variations of FAD, proposing that certain embedding models provide results that correlate well on a per-song basis with subjective evaluation criteria for audio and music quality \cite{gui2024adapting}.

In addition to the existing objective metrics, previous approaches use a range of subjective metrics for evaluation, two prominent metrics being Mean Opinion Score (MOS) and head-to-head comparisons (HTH). MOS is a commonly used metric in which listeners rate audio samples on a scale of 1 to 5 based on specific criteria. HTH comparisons of models are used to determine the winner based on specific criteria such as text alignment or music quality \cite{agostinelli2023musiclm, huang2023noise2music}. 

\section{Dataset}

A high-quality reference dataset is essential for evaluating music preference and text-audio alignment. This dataset should provide realistic music descriptions paired with corresponding audio tracks, enabling music generation with various models and serving as a reliable benchmark for both human and objective evaluation. We selected the MTG-Jamendo dataset~\cite{bogdanov2019mtg}, which contains 55k tracks. The dataset is annotated with 195 distinct tags across genre, instrument, and mood/theme categories. 

\label{sec:datasets}

\subsection{Tag Selection}

To ensure effective tag-based music descriptions based on the reference dataset, we select tags that are commonly used in practice and combinations of tags that have an appropriate length for describing the generated music. Additionally, we ensure that the tags offer sufficient diversity to allow for a meaningful comparison of the models' capabilities, particularly in terms of text-audio alignment. \par

We remove tags that are not commonly used in practice by filtering out all tags that do not appear in the FMA \cite{defferrard2016fma} or MagnaTagATune \cite{law2009evaluation} datasets. Additionally, we choose tag combinations of length three to ensure a consistent length that is descriptive enough and has enough unique tag combinations of that exact length in the MTG-Jamendo dataset. After these steps, we are left with 1,248 unique tag combinations with at least one track in the reference dataset. To ensure a diverse set of tag-based music descriptions, we ensure that no two tag-combinations have a CLAP embedding \cite{elizalde2024natural} with a cosine similarity value above a threshold of 0.1382. The threshold is selected with a binary search so that the final set of tag-based music descriptions is 500.

\subsection{Music Generation}

For each one of 12 music generation models we generate 500 music tracks with the selected prompts to create a dataset of 6,000 AI-generated music tracks. The models cover a diverse range of capabilities, enabling a comprehensive comparison between human and objective evaluations. For MusicGen~\cite{copet2024simple}, a transformer-based music generation model, we generate clips from the three checkpoints ``musicgen-small'', ``musicgen-medium'', and ``musicgen-large''. Additionally, we generate music with diffusion-based models Riffusion~\cite{forsgren2022riffusion}, AudioLDM 2~\cite{liu2023audioldm2} (``audioldm2-music'' and ``audioldm2-large'' checkpoints), Mustango \cite{melechovsky2023mustango} and Stable Audio~\cite{evans2024fast} (``Stable Audio AudioSparx 1.0'' and ``Stable Audio AudioSparx 2.0''). Furthermore, we evaluate two state-of-the-art commercial music generation models, Suno~\cite{suno} (Suno v3 and Suno v3.5) and Udio~\cite{udio}, which have recently gained attention for their ability to generate high-quality audio across a wide range of music styles and genres. \par

Given that many of these models were designed to generate shorter music clips without vocals or lyrics, we limit the generated track duration to 10 seconds and instrumental versions only.
For models such as Suno-v3, Suno-v3.5, and Udio, as well as tracks from the MTG-Jamendo dataset that tend to exceed 10 seconds, we select a 10-second segment that contains the highest average energy. This was done to ensure that we do not randomly select sections with silence.

\section{Human Evaluation}
\label{sec:prolificds}

\subsection{Survey Design}

We focus on two key metrics: text-audio alignment and human music preference. Text-audio alignment assesses how accurately a model can generate music that abides by a given textual input. 
In addition, we measure human music preference to determine which music generation methods yield the subjectively best results. We take inspiration from human evaluations of LLM chatbots \cite{chiang2024chatbot} and use a similar methodology and evaluation technique in our study.
To evaluate music preference and text-audio alignment, we design a survey using pairwise comparisons with binary preference choices between two music clips. Each survey question presents participants with two music tracks, each clipped to 10 seconds. For the evaluation of music preferences, participants were asked ``What music clip do you prefer?''. For text-audio alignment, participants were asked ``Which music clip is better described by'' followed by the combination of tags used to generate the track (in the case of the reference dataset, the original tagging was used). Respondents were limited to a binary selection of either ``Music Clip 1'' or ``Music Clip 2.''

For the human evaluation, we randomly select 100 tag combinations and the corresponding generated music for each music generation model, along with the baseline tracks from MTG-Jamendo. As each track for each text description is compared with every other track with that same description, the resulting survey comprises a total of 7,800 music preference questions and 7,800 text alignment questions.

We use the Prolific platform to run the survey.~\cite{prolific} Although previous music evaluation studies have often relied on Amazon Mechanical Turk,~\cite{mturk} recent research suggests that Prolific produces higher quality responses~\cite{douglas2023data}. We pre-filter the survey audience to include only fluent English speakers within the age range of 18 to 34 years who reported using music streaming services. Each participant completed a survey consisting of three music preference questions and three text alignment questions. The order of the questions was randomly shuffled for each participant. Additionally, we include an attention check in the form of a text alignment question featuring a high-quality track from MTG-Jamendo paired with static white noise.

\subsection{Survey Analysis}
\begin{figure}[t!]
\includegraphics[width=\linewidth]{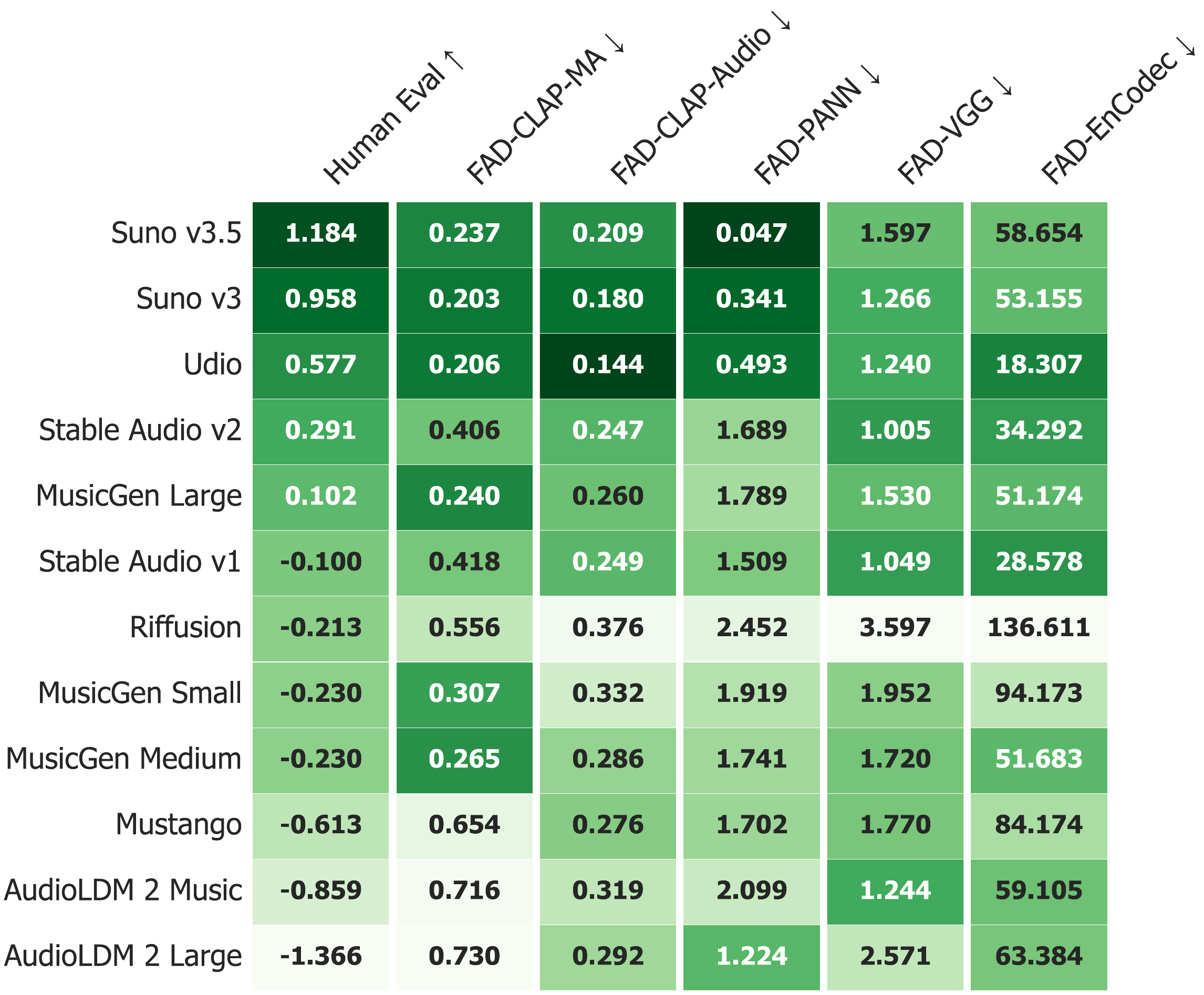}
\caption{Scores of the tested metrics for music preference estimation on the generated audio samples and Bradley-Terry parameters of our human evaluation. The cell colors indicate better (green) or worse (white) scores. FAD-PANN scores were multiplied by 1,000.}
\label{fig:human_eval_mp}
\end{figure}

We present bootstrapped Elo ratings for the music preference and text-audio alignment survey results in \cref{fig_wins_scatterplot}. 
The Elo ratings were initialized with a base rating of 1,000 and a K-factor of 8. With \(R_A\) and \(R_B\) as the current Elo ratings of model \(A\) and \(B\) and \(R'_A\) and \(R'_B\) the updated ratings, the Elo ratings update formula, for each pairwise comparison of two models, is computed as follows:

\begin{equation}
R'_A = R_A + K \cdot (S_A - \frac{1}{1 + 10^{(R_B - R_A)/400}})
\label{elo_update}
\end{equation}

We set \(S_A = 1\) if model \(A\) wins and \(S_A = 0\) if model \(A\) loses. The same formula applies to model \(B\), with the variables of \(A\) and \(B\) switched. 
We randomly shuffle the pairwise comparisons 10k times and report the mean Elo rating across the bootstrapping procedure for each model.

For music preference, commercial models like Suno v3.5, Suno v3, and Udio all outperformed MTG-Jamendo. Notably, Suno v3.5 achieved a significantly higher Elo than all other models. As expected, newer and larger versions of models generally performed better. For instance, Stable Audio v2 exhibited nearly a 5 percentage point increase over Stable Audio v1. Similarly, MusicGen Large outperformed MusicGen Medium and MusicGen Small. 
Suno v3.5 also obtained the best rating for the text-audio alignment rating with a considerable margin over the other models.

\section{Metric Comparison}
\label{sec:analysis}

\begin{figure}[t!]
\includegraphics[width=\linewidth]{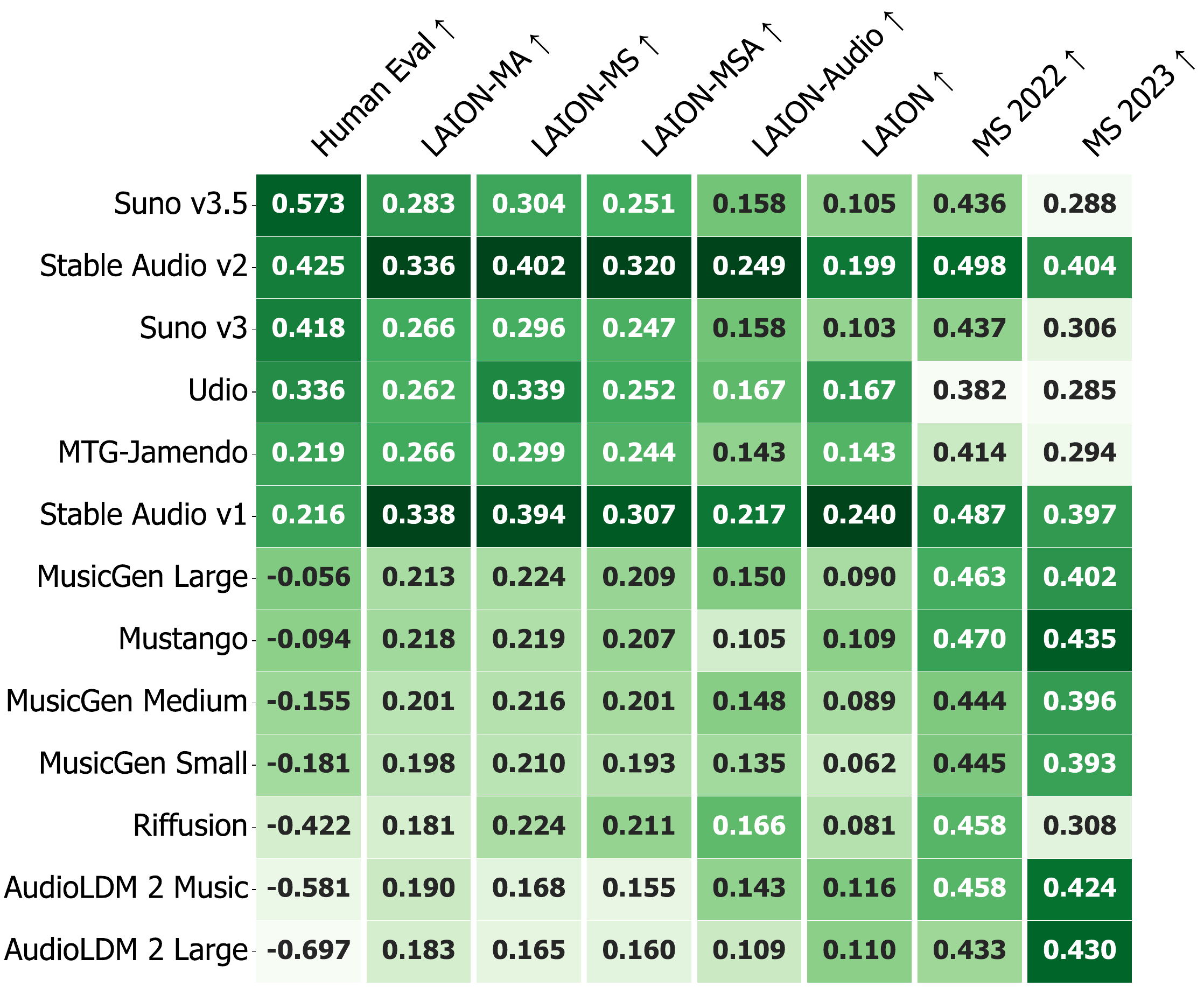}
\caption{Scores of the tested metrics for text-audio alignment on the generated audio samples and Bradley-Terry parameters of our human evaluation. The cell colors indicate better (green) or worse (white) scores. }
\label{fig:human_eval}
\end{figure}

\begin{figure}
\begin{minipage}[b]{.41\linewidth}
\centering
\includegraphics[width=\textwidth]{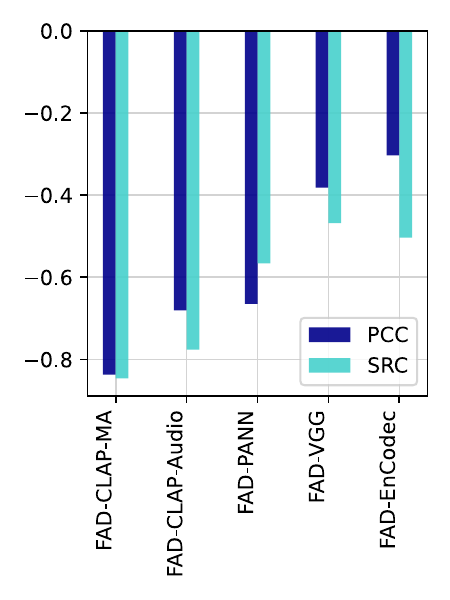}
\end{minipage}
\hfill
\begin{minipage}[b]{0.55\linewidth}
\centering
\includegraphics[width=\textwidth]{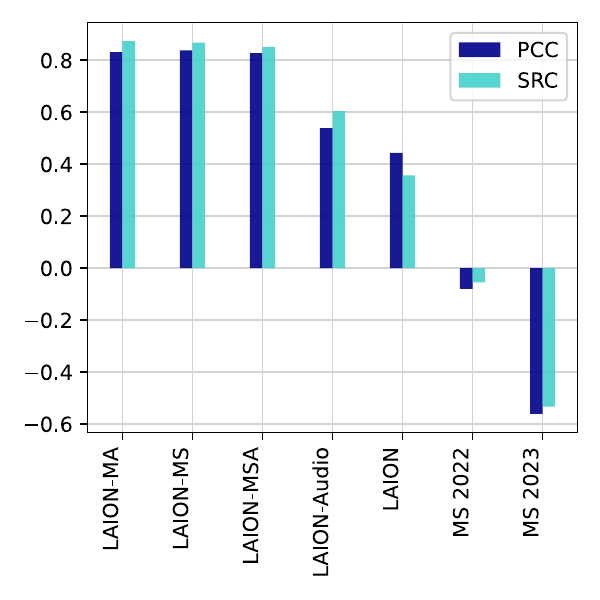}
\end{minipage}

\caption{Pearson correlation coefficient (PCC) and Spearman's rank correlation coefficient (SRC) between objective metrics and Bradley-Terry parameters of the human evaluation results. \textbf{(Left)} Music preference metric correlations (lower is better). \textbf{(Right)} Text-audio alignment metric correlations (higher is better).}
\label{fig_correlations}
\end{figure}

To compare subjective and objective metrics and thus measure how well the tested metrics align with human perception, we compute the Bradley-Terry parameters~\cite{maystre2015fast,bradley1952rank} to indicate the ``strength'' of the music generation models for both the pairwise comparisons of music preference and text-audio alignment.
We report the results for the objective metrics in \cref{fig:human_eval_mp} and \cref{fig:human_eval} and the correlation of the objective metrics with the Bradley-Terry parameters of the subjective evaluation in \cref{fig_correlations}, where we compute the Pearson correlation coefficient and Spearman's rank correlation coefficient. 

We use FAD with different embedding models to evaluate how objective metrics can approximate human music preference. We employ VGGish~\cite{hershey2017cnn} (FAD-VGG), and PANN~\cite{kong2020panns} (FAD-PANN), which are audio classification models. For CLAP, we use the audioset and music-audioset checkpoints with \texttt{630k-audioset-best} (FAD-CLAP-Audio) and \texttt{music\_audioset\_epoch\_15\_esc\_90.14} (FAD-CLAP-MA). 
Additionally, we use the 24 kHz mono version of EnCodec~\cite{defossez2022high} (FAD-EnCodec), an audio compression model. 
The MS-CLAP \cite{elizalde2023clap,elizalde2024natural} and LAION-CLAP \cite{wu2023large} models are used to evaluate text-audio alignment. Specifically, for LAION-CLAP, we analyze the checkpoints \texttt{630k-best} (LAION), \texttt{630k-audioset-best} (LAION-Audio), \texttt{music\_audioset\_epoch\_15\_esc\_90.14} (LAION-MA), \texttt{music\_speech\_epoch\_15\_esc\_89.25} (LAION-MS), and \texttt{music\_speech\_audioset\_epoch\_15\_esc\\\_89.98} (LAION-MSA). For the MS-CLAP models, we consider the ``2022'' (MS 2022) and ``2023'' (MS 2023) versions. For all CLAP and LAION-CLAP models we compute the mean cosine-similarity between the audio embeddings and the tag-based descriptions for each music generation model.

FAD-CLAP-MA demonstrates the best correlation with human perception of music quality in terms of both linear and rank correlation with human evaluation. This finding aligns with prior FAD results~\cite{gui2024adapting}, computed per song on a smaller set of music generation models. %
A closer look at ~\cref{fig:human_eval_mp} indicates that PANN is an exemplary embedding model for identifying high-quality music generation models such as Suno and Udio. We can also observe that all models tend to rate Riffusion worse than our human preference study suggests and usually rank it at the last place (except for FAD-CLAP-MA).

For text-audio alignment, CLAP models trained on music data (LAION-MA, LAION-MS, and LAION-MSA) exhibit the highest correlation with human ratings. Additionally, the strong correlations observed in Pearson's correlation coefficient, as well as Spearman's rank correlation coefficients, indicate that the cosine similarity values from those models demonstrate substantial linear and rank correlation with human judgments. Overall, the rankings for both music quality perception and text-audio alignment suggest that the LAION-MA checkpoint aligns best with human preferences and consistently outperforms others.

\section{Conclusions}
\label{sec:conclusions}

We present a comprehensive human study on the performance of current generative music models, an emerging field that lacked a comprehensive benchmark until now. We produce a corpus of music by selecting common tag combinations from MTG-Jamendo~\cite{bogdanov2019mtg} and utilizing a diverse range of both open-source and commercial music generation models. Through a large-scale human survey, we collect detailed feedback on human music preference and text-audio alignment, providing an unbiased ranking of the models. We find that the commercial model Suno \cite{suno} aligns exceptionally well with human evaluation, outperforming even the reference dataset MTG-Jamendo by considerable margins. Further, it enables us to benchmark and access the alignment of existing metrics with human perception. Among the metrics we tested, CLAP models \cite{wu2023large} trained on music data most accurately approximate human preferences, both when employed as embedding models for computing FAD scores and for approximating text-audio alignment. We make all associated artifacts publicly available (including human evaluations) to support future research and the development of better metrics.

\bibliographystyle{IEEEtran}
\bibliography{refs}

\end{document}